\documentclass{article}
\usepackage{iclr2027_conference,times}
\usepackage{amsmath,amssymb,mathtools}
\usepackage{graphicx,booktabs,tabularx,array,multirow}
\usepackage{hyperref,url,xcolor}
\usepackage{enumitem}
\usepackage{placeins}
\usepackage{capt-of}
\hypersetup{colorlinks=true,linkcolor=blue!45!black,citecolor=blue!45!black,urlcolor=blue!45!black}
\newcommand{\E}{\mathbb{E}}

\newcommand{\tr}{\operatorname{tr}}
\newcommand{\norm}[1]{\left\lVert #1\right\rVert}

\newcommand{\methodfull}{REad-geometry CAlibration with dynamic Prefixes}
\newcommand{\method}{RECAP}

\newcolumntype{Y}{>{\raggedright\arraybackslash}X}
\title{RECAP: Calibrating Low-Bit Reasoning\\Models on the History They Read}
\author{Chi Ruan$^{1}$,\enspace Yujie Chen$^{2}$,\enspace Zixiang Zhao$^{2}$,\enspace Jingcai Guo$^{3}$,\enspace Haotong Qin$^{3}$\thanks{Corresponding author: \href{mailto:haotong.qin@polyu.edu.hk}{haotong.qin@polyu.edu.hk}.}\\
\normalfont $^{1}$SpinEntropy,\enspace $^{2}$ETH Zurich,\enspace $^{3}$PolyU}
\hypersetup{pdfauthor={Chi Ruan, Yujie Chen, Zixiang Zhao, Jingcai Guo, Haotong Qin},
pdftitle={RECAP: Calibrating Low-Bit Reasoning Models on the History They Read}}
\iclrfinalcopy
\begin{document}
\maketitle
\lhead{Under review as a conference paper at ICLR 2027}
\begin{abstract}
Low-bit quantization reduces the cost of serving reasoning models but can substantially degrade long chains of thought. Each step reads cached keys and values spanning thousands of tokens, allowing quantization errors to accumulate along the trajectory. Standard calibration operates on short segments that omit the earlier context needed by later reasoning steps. Even when this context is retained, cache reconstruction error alone does not capture how quantization errors affect subsequent attention. We present \method{}, a post-training quantization method that calibrates reasoning models on the history they read. Dynamic-prefix calibration separates long historical context from short-block reconstruction. It regenerates prefix keys and values with the current quantizer and rotates the active block to supervise every position along the trajectory. Prefix read geometry further supervises historical cache errors according to their effects on attention outputs, weighting key and value errors asymmetrically. With both components applied only during calibration, \method{} outperforms the baseline across multiple reasoning models and quantization settings. Under W4A4KV4, \method{} improves accuracy over the baseline by 13.3 percentage points on MATH500 with DeepSeek-R1-Distill-Qwen-1.5B and 15.8 points on AIME-120 with Qwen3-4B. These results highlight the importance of preserving historical dependencies during calibration for low-bit reasoning.
Project repository: \mbox{\url{https://github.com/AI-Efficiency/RECAP}}.
\end{abstract}
\suppressfloats[t]
\begin{figure}[t]
\centering
\includegraphics[width=\linewidth]{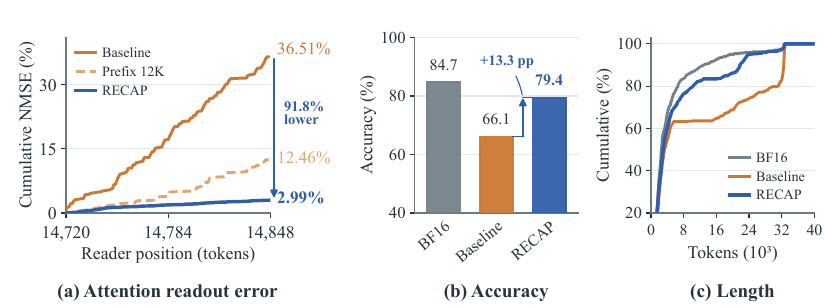}
\caption{Attention error and reasoning performance of DeepSeek-R1-Distill-Qwen-1.5B. (a) Cumulative normalized attention error with a 12K-token INT4 KV prefix. (b,c) MATH500 accuracy and re-tokenized output-length distributions under W4A4KV4, with BF16 as the reference.}
\label{fig:calibration-overview}
\end{figure}

\section{Introduction}
Reasoning language models solve complex problems by generating extended sequences of intermediate steps~\citep{wei2022chain,deepseek2025r1,qwen2025original4b}. These capabilities come with substantial inference costs, making low-bit quantization an attractive approach to deployment~\citep{frantar2023gptq,lin2024awq,xiao2023smoothquant}. However, low-bit quantization can substantially degrade reasoning performance~\citep{liu2025quantization,lv2026reasoningqat,lee2026reqat}. Long-chain reasoning relies on repeated access to a growing history of intermediate states. Quantization errors in cached keys and values can distort attention outputs at later reasoning steps. As shown in Figure~\ref{fig:calibration-overview}b,c, quantization can reduce answer accuracy while increasing generation length. This motivates calibration that preserves the historical dependencies underlying long-chain reasoning.

Existing post-training quantization methods seek to preserve model accuracy by improving the fidelity of quantized representations. QuaRot~\citep{ashkboos2024quarot} suppresses activation outliers through rotations, while FlatQuant~\citep{sun2025flatquant} learns affine transformations through layer-output reconstruction. However, for reasoning models with low-bit weights, activations, and KV caches, calibration effectiveness also depends on how well it captures the role of earlier quantized states in later computation. Layer-output reconstruction captures these interactions within the supplied historical context and at the positions being reconstructed. Improving representation fidelity within this scope does not by itself ensure that longer historical dependencies are preserved across the trajectory. These considerations motivate examining which history calibration exposes, which reader positions it supervises, and how historical errors affect subsequent attention.

Our analysis identifies three challenges in preserving these historical dependencies. First, the cache-importance scores we tested transfer poorly across reasoning trajectories, limiting their usefulness for selecting which states to preserve. Calibration should therefore retain the historical context itself rather than a selected subset. Retaining this context raises a second challenge, which is to balance historical context, position coverage, and computational cost. Short windows omit relevant history, reconstructing only the final segment leaves earlier positions unsupervised, and reconstructing the full sequence at every update is costly. Third, even with adequate context and coverage, cache errors of similar magnitude can distort attention to different degrees, so raw cache reconstruction error remains an incomplete measure of their effect on later computation.

To address these challenges, we introduce \method{} (\methodfull{}), a post-training quantization framework that calibrates reasoning models on the history they read. \emph{Dynamic-prefix calibration} addresses history and position coverage by allowing each short reconstruction block to read a long prefix regenerated with the current quantizer. Rotating the active block along the trajectory extends supervision to every position without reconstructing the entire sequence at each update. With this coverage in place, \emph{prefix read geometry} further accounts for how errors in the retained history affect subsequent attention. It penalizes historical key and value errors according to their effects on attention outputs, complementing block reconstruction with direct supervision of the history being read.

Together, these components preserve historical dependencies during calibration without adding inference-time token selection or precision allocation. We evaluate \method{} across multiple reasoning models and quantization settings on mathematical reasoning, scientific question answering, and code generation. \method{} consistently matches or improves on the baseline. Under W4A4KV4, accuracy gains reach up to 13.3 percentage points on MATH500 and 15.8 points on AIME-120. Our contributions are as follows:
\begin{itemize}[leftmargin=*,itemsep=0pt,topsep=2pt]
\item \textbf{Motivation analysis.} We show that calibrating low-bit reasoning models is limited by historical dependencies: cache-importance scores transfer across trajectories at near-chance level, short-window and final-segment calibration each fail on the positions they leave uncovered, and cache errors of equal size affect later attention unequally.
\item \textbf{Dynamic-prefix calibration.} We decouple the context used for calibration from the positions being reconstructed. At each update, the historical keys and values are regenerated with the current quantizer, and only a short block is reconstructed; this block is shifted along the trajectory across updates. Calibration therefore observes the long quantized history encountered at inference, and every position is supervised at the cost of short-block reconstruction.
\item \textbf{Prefix read geometry.} We introduce an auxiliary cache objective derived from the local sensitivity of the attention output. Key errors are weighted by their projection onto query directions, and value errors by the attention they receive. Historical errors are thus penalized according to their influence on subsequent attention, which reconstruction error alone does not capture.
\end{itemize}

\begin{figure}[t]
\centering
\includegraphics[width=\linewidth]{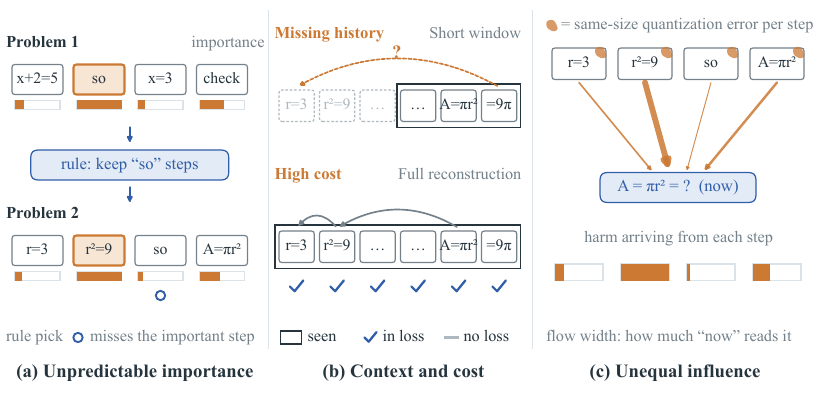}
\caption[Quantization sensitivity and calibration coverage.]{Quantization sensitivity and calibration coverage. (a) A rule that finds the important cached step in one problem misses the important step in another. (b) Short calibration windows omit historical dependencies, while reconstructing every position in a long sequence incurs a higher computational cost per update. (c) The current step reads its cached history with unequal weight, so the same-size error on different steps has unequal influence.
}
\label{fig:readback-motivation}
\end{figure}

\section{Motivation}
\label{sec:motivation}
We first review reconstruction-based quantization and cached attention. We then examine three calibration challenges in reasoning models: limited transfer of importance scores, incomplete history and position coverage, and the mismatch between cache error and attention distortion.

\subsection{Preliminaries}
\label{sec:preliminaries}

Low-bit quantization represents weights, activations, and cached keys and values with reduced precision. Reconstruction-based post-training quantization calibrates layer-specific parameters $\theta_\ell$ on reference hidden states $H^\ell$ by minimizing
\begin{equation}
\min_{\theta_\ell}\;\E_{H^\ell}\Bigl[\bigl\lVert
f_{\theta_\ell}^{\ell}(H^\ell)-f_{\mathrm{FP}}^{\ell}(H^\ell)\bigr\rVert_F^2\Bigr],
\label{eq:prelim-reconstruction}
\end{equation}
where $f_{\mathrm{FP}}^\ell$ and $f_{\theta_\ell}^\ell$ denote the FP and quantized layers~\citep{shao2024omniquant,sun2025flatquant}. The calibration sequences determine both the available context and the positions included in this objective.

\begin{figure}[t]
\centering
\includegraphics[width=\linewidth]{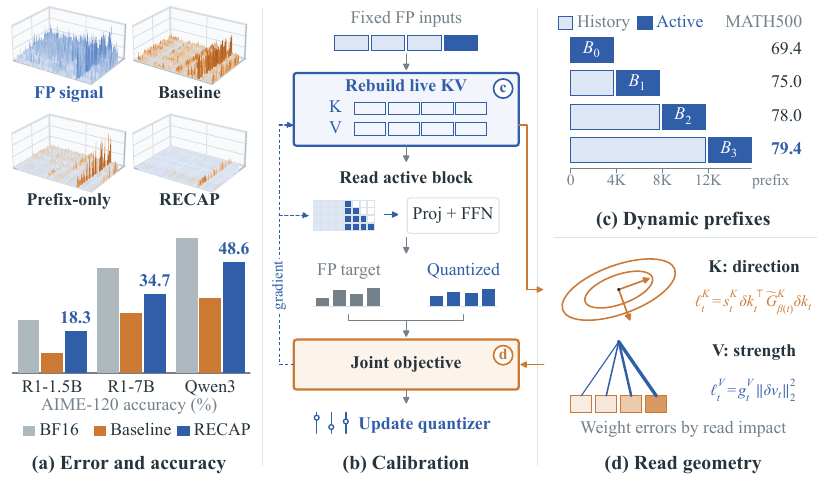}
\caption{RECAP overview.
(a) Full-precision attention magnitudes alongside quantization errors for the baseline, prefix-only calibration, and RECAP in one reader window; bars give W4A4KV4 AIME-120 accuracy from Table~\ref{tab:comparison-main} on R1-1.5B, R1-7B, and Qwen3-4B.
(b) The calibration workflow rebuilds historical keys and values with the current quantizer and combines block reconstruction with a read-geometry penalty.
(c) Dynamic prefixes retain historical context as the active block rotates along the reasoning trajectory.
(d) Read geometry weights key errors by direction and value errors by read strength, as in Equation~\ref{eq:aux}; prefix and head indices are omitted.}
\label{fig:method}
\end{figure}

Within one attention head, query $q_s$ reads causally available keys $k_t$ and values $v_t$ through
\begin{equation}
a_{st}=\frac{\exp\bigl(q_s^\top k_t/\sqrt{d_h}\bigr)}
{\sum_{t'\leq s}\exp\bigl(q_s^\top k_{t'}/\sqrt{d_h}\bigr)},
\qquad
y_s=\sum_{t\leq s}a_{st}\,v_t,
\label{eq:prelim-attention}
\end{equation}
where $d_h$ is the head dimension. During autoregressive reasoning, earlier K/V remain in the cache and can influence later steps whenever they are read.

\subsection{Quantization Sensitivity and Calibration Coverage}
\textbf{Limited transfer of importance scores.} As illustrated in Figure~\ref{fig:readback-motivation}(a), an importance score that works for one reasoning problem may not identify the important cached states in another. We restore individual quantized cache units to full precision and measure how much each repair reduces the teacher-forced loss of the final answer. Across four reasoning trajectories, scores fitted on three trajectories and evaluated on the held-out one rank the most beneficial 10\% of repairs close to chance, with AUC between 0.43 and 0.50. The importance patterns observed in one reasoning trajectory, therefore, do not consistently carry over to another. This suggests that the tested scoring rules provide limited guidance for selecting important cached states across problems.

\textbf{Historical context, position coverage, and cost.} Figure~\ref{fig:readback-motivation}(b) illustrates a trade-off in calibrating long reasoning sequences. Short windows can omit earlier premises that later steps depend on, preventing calibration from capturing how those steps use the preceding history. Extending the window restores access to more of this context, but reconstructing every position requires more computation per update as the sequence grows longer. Long reasoning trajectories, therefore, make full-sequence reconstruction costly. Supplying a long prefix while reconstructing only the final segment restores history, but leaves earlier query positions outside the objective. In teacher-forced diagnostics over an 8K context, a model calibrated on 4K windows diverges from the FP model late in the sequence, with KL rising from 0.14 nats/token in the first 1K positions to 7.33 in the last 1K. A model that reconstructs only positions 4K--8K with a 4K prefix fails in the opposite region, reaching 6.58 in the first 1K and 0.17 in the last. The challenge is to preserve the historical information needed by later steps and supervise every position without reconstructing the entire sequence in every update.

\textbf{Unequal influence of cache errors.} Figure~\ref{fig:readback-motivation}(c) illustrates how quantization errors of the same size can have different effects on the current reasoning step. The current query reads the cached history with unequal attention weights. In the example, it relies more heavily on the step $r^2=9$ than on the connecting word ``so'', so errors at these positions can contribute differently to the attention output. Measuring cache error alone does not capture this difference~\citep{zhou2026oscar}. Although layer-output reconstruction captures these effects at supervised positions, an additional cache loss should also reflect how later queries read the history.

These observations suggest three requirements for calibration. It should retain the historical context that later steps read, extend supervision across the trajectory without reconstructing the full sequence at every update, and account for how cache errors affect subsequent attention outputs.

\FloatBarrier

\section{\method{}}
\label{sec:method}
This section presents \method{}, with its overall design illustrated in Figure~\ref{fig:method}. We first establish the calibration setting by describing the reference reasoning trajectories and the layer reconstruction objective. We then introduce dynamic prefixes to preserve historical context during reconstruction, followed by prefix read geometry to account for how cache errors affect later attention outputs.

\subsection{Calibration Formulation}
We build a calibration set using reasoning trajectories generated by the original full-precision model~\citep{williams2025selfcalibration}:
\begin{equation}
\mathcal D_{\mathrm{self}}=\{(x_i,r_i)\}_{i=1}^{N}.
\label{eq:self-data}
\end{equation}
Here, $x_i$ is an input problem and $r_i$ is the model's generated reasoning and answer. We tokenize them together, preserve their causal order and original positions, and cache the full-precision layer inputs $H_i^\ell$ and outputs $Y_i^\ell=f_{\mathrm{FP}}^\ell(H_i^\ell)$. Calibration optimizes FlatQuant's learnable transformations, diagonal scales, and clipping parameters~\citep{sun2025flatquant}, while pretrained weights remain fixed.

For quantization parameters $\theta_\ell$, let $\widehat Y_i^\ell=f_{\theta_\ell}^{\ell}(H_i^\ell)$ denote the quantized layer output. Full-sequence reconstruction minimizes its mean squared error relative to the full-precision output:
\begin{equation}
\mathcal L_{\ell}^{\mathrm{full}}(\theta_\ell)
=\frac{1}{N}\sum_{i=1}^{N}\mathcal L_{i,\ell}^{\mathrm{full}}(\theta_\ell),
\qquad
\mathcal L_{i,\ell}^{\mathrm{full}}(\theta_\ell)
=\frac{1}{T_i d}\bigl\lVert\widehat Y_i^\ell-Y_i^\ell\bigr\rVert_F^2,
\label{eq:full}
\end{equation}
where $T_i$ is the sequence length and $d$ is the layer output dimension. Calibration is performed layerwise with teacher forcing, keeping the full-precision inputs and reference outputs fixed while optimizing $\theta_\ell$.

\subsection{Dynamic Prefixes}
Each calibration step reconstructs one block while retaining its preceding K/V. We partition a trajectory of length $T$ into blocks $B_j=[jB,(j+1)B)$ of length $B$, and omit the trajectory index $i$ below, so $\mathcal L_\ell^{\mathrm{full}}$ denotes the per-trajectory loss $\mathcal L_{i,\ell}^{\mathrm{full}}$. For an active block beginning at $P=jB$, we generate its prefix from the fixed FP layer inputs:
\begin{equation}
C_{<P}(\theta_\ell)=\operatorname{KV}_{\theta_\ell}\bigl(H^{\ell}_{<P}\bigr).
\label{eq:cache}
\end{equation}
The KV operator applies the current layer normalization, projection, activation, and cache quantization rules with the original positional transformations. We rebuild this prefix at every microstep to keep it consistent with the changing quantization parameters.

Queries in the active block attend to the prefix and causally available K/V within the block. We reconstruct only the active block's outputs:
\begin{equation}
\mathcal L_\ell^{\mathrm{block}}(\theta_\ell;j)
=\frac{1}{Bd}\sum_{t\in B_j}
\norm{\widehat Y^{\ell}_{t}\bigl(H^{\ell}_{B_j},C_{<P}(\theta_\ell)\bigr)-Y^{\ell}_{t}}_2^2.
\label{eq:block}
\end{equation}
Gradients flow through both the active block and its prefix K/V. The block loss can therefore update the parameters that produce historical representations without computing the prefix's query, attention, or feed-forward outputs.

To supervise every position, we rotate the active block across the $J=T/B$ blocks: at epoch $e$, the sample with stable index $i$ uses block $j=(e+i)\bmod J$, so each sample visits all $J$ blocks once in every $J$ epochs. The first block has an empty prefix and uses an ordinary causal forward pass.

At fixed parameters, the block losses partition the full-sequence objective. When the prefix is generated with the same transformations, quantization rules, and original positions as the full forward pass, each active-block output equals the corresponding full-sequence output, so
\begin{equation}
\frac{1}{J}\sum_{j=0}^{J-1}\mathcal L_\ell^{\mathrm{block}}(\theta_\ell;j)=\mathcal L_\ell^{\mathrm{full}}(\theta_\ell).
\label{eq:decomposition}
\end{equation}
A uniformly sampled block, therefore, gives an unbiased estimate of Equation~\ref{eq:full} and, because gradients flow through the prefix K/V, of its gradient. Each update computes queries, attention outputs, and feed-forward outputs only for the active block. The identity requires token-wise quantization statistics and the same causal support; parameter updates between blocks mean that a full rotation is not identical to full-sequence optimization.

\subsection{Prefix Read Geometry}
\label{sec:curvature}
The block loss in Equation~\ref{eq:block} measures errors at active outputs. Although this loss already reaches the prefix K/V through gradients, that signal is indirect: it is mixed with weight and activation errors at the active outputs and reflects only the queries of the current block. We therefore add a \emph{prefix read geometry} loss that supervises each historical entry directly, weighting its error by how strongly later queries read it. Its metric is collected once from the original FP target model and frozen before layerwise calibration.

\paragraph{Attention sensitivity.}
Suppressing layer and head indices, let $q_s,k_t,v_t\in\mathbb R^{d_h}$ be FP reference attention tensors in physical coordinates: Q/K are post-RoPE, and reversible cache transformations have been removed. Let $a_{st}$ and $y_s$ denote the attention weights and head output in Equation~\ref{eq:prelim-attention}. For fixed queries and perturbations only at prefix positions $t<P$, the first-order change is
\begin{equation}
\delta y_s=\sum_{t<P}a_{st}\delta v_t
+\sum_{t<P}\frac{a_{st}}{\sqrt{d_h}}
(v_t-y_s)q_s^\top\delta k_t.
\label{eq:read-jacobian}
\end{equation}
Thus a key error acts through both its alignment with a query and the value contrast $v_t-y_s$. A value error enters directly, scaled by $a_{st}$.

For one-half the mean squared Euclidean distortion at the attention head output, the local generalized Gauss--Newton (GGN) is the attention Jacobian's Gram matrix. Retaining its individual K and V diagonal blocks gives
\begin{equation}
G^K_{t,P}=\E_{s\in\mathcal Q_P}
\left[\frac{a_{st}^2\norm{v_t-y_s}_2^2}{d_h}\,q_sq_s^\top\right],
\qquad
g^V_{t,P}=\E_{s\in\mathcal Q_P}\bigl[a_{st}^2\bigr],
\label{eq:read-blocks}
\end{equation}
where $\mathcal Q_P$ contains assistant query positions in the active block $[P,P+B)$, and the value block is $g^V_{t,P}I$. For grouped-query attention (GQA), we sum over all query heads sharing a KV head. The metric describes local head-output distortion before the output projection and omits cross-position and K--V cross terms.

\paragraph{Calibration objective.}
We retain per-entry key strength $s^K_{t,P}=\tr(G^K_{t,P})$ and value strength $g^V_{t,P}$. Unit-trace key matrices are pooled across fit trajectories and prefix conditions within each position bucket and KV head. Their shared approximation $\widetilde G^K_{\beta(t)}$, where $\beta(t)$ denotes the bucket containing position $t$, combines low-rank principal directions with an isotropic discarded-spectrum residual. This is a pooled approximation to Equation~\ref{eq:read-blocks}, not a full matrix per entry.

To compare quantized and FP caches, first undo the current learned transformations. In row-tensor notation, define $\Delta K=\widehat K'_{\theta}T_{K,\theta}^{-1}-K_{\mathrm{FP}}$ and $\Delta V=\widehat V'_{\theta}T_{V,\theta}^{-1}-V_{\mathrm{FP}}$. FP targets use the same fixed normalized layer inputs and original positions. The true inverses retain gradients. This avoids penalizing a change of cache basis as if it were quantization error.

Writing $\delta k_{t,h},\delta v_{t,h}$ for the corresponding error vectors at KV head $h$, we define the read-geometry penalty and combine it with block reconstruction:
\begin{equation}
\begin{aligned}
\mathcal R_\ell&=\frac{1}{H_{\mathrm{KV}}P}\sum_{h=1}^{H_{\mathrm{KV}}}\sum_{t<P}
\Bigl[s^K_{t,P,h}\,\delta k_{t,h}^{\top}\widetilde G^K_{\beta(t),h}\,\delta k_{t,h}
+g^V_{t,P,h}\norm{\delta v_{t,h}}_2^2\Bigr],\\
\mathcal L_\ell&=\mathcal L_\ell^{\mathrm{block}}(\theta_\ell;j)+\alpha_\ell\,\mathcal R_\ell.
\end{aligned}
\label{eq:aux}
\end{equation}
Here, $\alpha_\ell$ is a fixed layer-specific coefficient. For $P=0$, $\mathcal R_\ell$ is defined to be zero. After calibration, neither the reference statistics nor the auxiliary loss is needed for inference.

\clearpage
\FloatBarrier
\section{Experiments}
\label{sec:experiments}
We first compare a range of models and quantization methods across diverse benchmarks, then examine paired generation behavior. We assess both task accuracy and generation length to characterize the effects of quantization on reasoning. Ablation studies assess historical-prefix length and the contribution of read geometry beyond MSE.

\subsection{Setup}
\label{sec:setup}
\paragraph{Models and benchmarks.}
We compare DeepSeek-R1-Distill-Qwen-1.5B/7B and Qwen3-4B, using BF16 as the unquantized reference and QuaRot and FlatQuant as quantization baselines. We evaluate MATH500~\citep{hendrycks2021math,lightman2023verify}, AIME-120, GSM8K~\citep{cobbe2021gsm8k}, GPQA-Diamond~\citep{rein2023gpqa}, and LiveCodeBench~\citep{jain2024livecodebench}, covering mathematical reasoning, scientific question answering, and code generation. We report task scores throughout and mean generation length in the MATH500 and AIME-120 ablations.

\paragraph{Quantization and calibration.}
We evaluate W4A4KV4 and W4A8KV4 with symmetric per-channel weights, asymmetric per-token activations, asymmetric KV groups of 128, and 16-bit queries. Calibration uses target-model FP trajectories; the 1.5B ablation pool contains 1,000 NuminaMath-1.5 traces~\citep{li2024numinamath}. The default uses 64 fit trajectories, a 12K maximum prefix, a rotating 4K block (16K total; 1K = 1,024 tokens), and 32 epochs (512 updates/layer). AdamW uses base learning rate $5\times10^{-3}$, weight decay 0.01, and cosine decay. We fix $\alpha_\ell$ at 0.1 times the lower median block-to-auxiliary gradient-norm ratio, computed at initialization over the first eight fit samples and all nonempty prefixes. The auxiliary multiplier is zero for the first 10\% of updates, rises linearly to one over the next 10\%, and then stays at one.

\paragraph{Evaluation protocol.}
Local generation uses temperature 0.6, top-$p$ 0.95, and a 32,768-token output limit. AIME-120 combines 90 problems from 2022--2024 and 30 from 2025. Scores are percentages, averaged equally across the five benchmarks. In the ablations, MATH500 and AIME-120 scores are averaged over three evaluations per configuration.

\subsection{Main Results}
\label{sec:comparisons}
\label{sec:main-comparisons}
Table~\ref{tab:comparison-main} summarizes the three-model comparison under both quantization settings. Published R1 baselines follow \citet[Table 1]{liu2025quantization}, and Qwen3-4B BF16 and W4A4KV4 FlatQuant follow \citet[Table 9]{lv2026reasoningqat}; the remaining entries are our full-dataset evaluations, averaged over three runs. The released 7B QuaRot K4 checkpoints degenerate during generation; we report their measured low scores.

\begin{table}[!t]
\caption{Comparison of quantization methods on three reasoning models under W4A4KV4 and W4A8KV4. Scores are percentages, with BF16 (gray) as the reference. MATH, AIME, GSM, GPQA, and LCB denote MATH500, AIME-120, GSM8K, GPQA-Diamond, and LiveCodeBench; Avg.\ is their unweighted mean. Bold marks the best quantized method per model and setting. R1-1.5B/7B: DeepSeek-R1-Distill-Qwen-1.5B/7B.}
\label{tab:comparison-main}
\label{tab:comparison-w4a8kv4}
\centering\footnotesize\setlength{\tabcolsep}{2.1pt}
\begin{tabular}{@{}llrrrrrrrrrrrr@{}}
\toprule
& & \multicolumn{6}{c}{W4A4KV4} & \multicolumn{6}{c}{W4A8KV4} \\
\cmidrule(lr){3-8}\cmidrule(l){9-14}
Model & Method & MATH & AIME & GSM & GPQA & LCB & Avg. & MATH & AIME & GSM & GPQA & LCB & Avg. \\
\midrule
R1-1.5B & BF16 & \textcolor{black!55}{84.7} & \textcolor{black!55}{23.3} & \textcolor{black!55}{84.5} & \textcolor{black!55}{36.2} & \textcolor{black!55}{16.4} & \textcolor{black!55}{49.0} & \textcolor{black!55}{84.7} & \textcolor{black!55}{23.3} & \textcolor{black!55}{84.5} & \textcolor{black!55}{36.2} & \textcolor{black!55}{16.4} & \textcolor{black!55}{49.0} \\
\cmidrule(l){2-14}
 & QuaRot & 1.4 & 0.0 & 0.9 & 12.0 & 0.0 & 2.9 & 0.6 & 0.0 & 0.7 & 22.2 & 0.0 & 4.7 \\
 & FlatQuant & 66.1 & 8.6 & 76.8 & 32.0 & 7.5 & 38.2 & 67.1 & 9.6 & 80.4 & 32.8 & 7.8 & 39.5 \\
 & \method{} & \textbf{79.4} & \textbf{18.3} & \textbf{79.9} & \textbf{32.8} & \textbf{12.1} & \textbf{44.5} & \textbf{82.1} & \textbf{22.1} & \textbf{80.9} & \textbf{34.3} & \textbf{11.9} & \textbf{46.3} \\
\midrule
R1-7B & BF16 & \textcolor{black!55}{93.9} & \textcolor{black!55}{46.1} & \textcolor{black!55}{91.2} & \textcolor{black!55}{51.0} & \textcolor{black!55}{36.7} & \textcolor{black!55}{63.8} & \textcolor{black!55}{93.9} & \textcolor{black!55}{46.1} & \textcolor{black!55}{91.2} & \textcolor{black!55}{51.0} & \textcolor{black!55}{36.7} & \textcolor{black!55}{63.8} \\
\cmidrule(l){2-14}
 & QuaRot & 1.4 & 0.0 & 0.4 & 24.6 & 0.0 & 5.3 & 1.0 & 0.0 & 0.6 & 22.4 & 0.0 & 4.8 \\
 & FlatQuant & 84.1 & 26.1 & \textbf{90.9} & 46.6 & 12.4 & 52.0 & 88.4 & 35.8 & 90.0 & 48.0 & 23.1 & 57.1 \\
 & \method{} & \textbf{91.0} & \textbf{34.7} & \textbf{90.9} & \textbf{48.7} & \textbf{18.8} & \textbf{56.8} & \textbf{93.0} & \textbf{43.3} & \textbf{90.1} & \textbf{50.0} & \textbf{26.9} & \textbf{60.7} \\
\midrule
Qwen3-4B & BF16 & \textcolor{black!55}{95.3} & \textcolor{black!55}{58.9} & \textcolor{black!55}{94.5} & \textcolor{black!55}{56.1} & \textcolor{black!55}{48.4} & \textcolor{black!55}{70.6} & \textcolor{black!55}{95.3} & \textcolor{black!55}{58.9} & \textcolor{black!55}{94.5} & \textcolor{black!55}{56.1} & \textcolor{black!55}{48.4} & \textcolor{black!55}{70.6} \\
\cmidrule(l){2-14}
 & QuaRot & 89.8 & 29.2 & 91.8 & 45.5 & 23.1 & 55.9 & 95.0 & 56.7 & \textbf{93.5} & 43.4 & 42.5 & 66.2 \\
 & FlatQuant & 89.9 & 32.8 & 92.1 & 47.5 & 29.1 & 58.3 & 94.4 & 57.5 & 92.8 & 50.0 & 45.5 & 68.0 \\
 & \method{} & \textbf{94.2} & \textbf{48.6} & \textbf{92.8} & \textbf{49.0} & \textbf{37.4} & \textbf{64.4} & \textbf{95.2} & \textbf{58.3} & \textbf{93.5} & \textbf{50.5} & \textbf{47.8} & \textbf{69.1} \\

\bottomrule
\end{tabular}
\end{table}

\paragraph{Consistent gains over FlatQuant.}
\method{} matches or exceeds FlatQuant on every model, bit setting, and benchmark in Table~\ref{tab:comparison-main}; the only tie is R1-7B on GSM8K under W4A4KV4 (90.9\%). Under W4A4KV4, its average score is 6.3, 4.8, and 6.1 points higher than FlatQuant on R1-1.5B, R1-7B, and Qwen3-4B, respectively; under W4A8KV4, the corresponding gains are 6.8, 3.6, and 1.1 points. Relative to BF16, \method{} retains 89.0--91.2\% of the average score under W4A4KV4, compared with 78.0--82.6\% for FlatQuant, closing 41--58\% of FlatQuant's remaining gap. Across all six model--precision combinations, \method{} achieves the highest average score among the quantized methods.

\paragraph{Gains concentrate on long multi-step reasoning.}
The largest improvements appear on AIME-120, MATH500, and LiveCodeBench, which require long reasoning chains or code generation. Under W4A4KV4, AIME-120 improves by 9.7, 8.6, and 15.8 points across the three models, and LiveCodeBench by 4.6, 6.4, and 8.3 points. On Qwen3-4B, the AIME-120 improvement recovers 60.5\% of FlatQuant's deficit to BF16. In contrast, the gains on GSM8K and GPQA-Diamond never exceed 3.1 and 2.1 points in either setting. This pattern is consistent with our motivation that errors in historical K/V matter most when later steps repeatedly read a long history, although the table alone does not isolate this mechanism.

\paragraph{Gains are largest where the baseline degrades most.}
FlatQuant degrades most on R1-1.5B, retaining 78.0\% and 80.6\% of the BF16 average under W4A4KV4 and W4A8KV4. This model also shows the largest average gains, 6.3 and 6.8 points, including 13.3 and 15.0 points on MATH500. When FlatQuant is already close to BF16, as for Qwen3-4B under W4A8KV4 (96.3\%), the gain narrows to 1.1 points. QuaRot collapses on both R1 models, with averages of at most 5.3\%, but remains competitive on Qwen3-4B, so the difficulty of low-bit quantization differs markedly across model families.

\FloatBarrier

\subsection{Ablation studies}
\label{sec:ablations}
\label{sec:prefix-selection}
\label{sec:p2-results}
\paragraph{Effect of prefix length.}
Table~\ref{tab:ablation-summary} compares 8K and 12K prefixes with a 4K active block on R1-1.5B under W4A4KV4. This increases total calibration coverage from 12K to 16K while retaining short-block reconstruction, so later query blocks can access a longer quantized history. For \method{}, the 12K configuration improves reported MATH500 and AIME-120 accuracy by 1.2 and 1.6 percentage points, respectively, while reducing mean generation length by 14.3\% and 10.8\%. Block MSE follows the same trend. These results favor the 12K configuration, although differences in calibration cohorts and update budgets limit attribution to prefix length alone.

\paragraph{Effect of read geometry.}
Block MSE already retains the prefix; read geometry adds direct supervision of historical K/V according to their effects on subsequent attention outputs. Key errors alter attention weights, whereas value errors perturb the content being aggregated. The reported read-geometry runs use statistics collected from the original FP model. Across both prefix lengths, \method{} improves reported accuracy by 1.0--1.6 points on MATH500 and 2.3--3.3 points on AIME-120, with 4.8--11.8\% fewer generated tokens. The gains therefore combine higher accuracy with shorter generation. These point estimates do not establish statistical superiority, and decoding differences further limit causal attribution at 12K.

\paragraph{Generation length and correctness.}
\label{sec:generation-behavior}
Figure~\ref{fig:generation-behavior} compares BF16 and FlatQuant on the same 500 MATH500 questions. Lengths are re-tokenized from saved outputs, with the dashed line marking the nominal 32K budget. Of the 101 questions that change from correct to incorrect, 96 produce longer quantized outputs, often approaching the generation limit. This association shows why generation length must be read together with correctness: long outputs can still end in wrong answers. Accordingly, the shorter generations in Table~\ref{tab:ablation-summary} are evaluated alongside their accuracy gains, without attributing errors to length alone.

\clearpage

\begin{table}[!t]
\centering
\makeatletter
\long\def\@makecaption#1#2{%
  \vskip 3pt
  {\fontsize{9}{10.5}\selectfont\raggedright #1: #2\par}}
\makeatother
\begin{minipage}[t]{0.45\linewidth}
\vspace{0pt}
\centering\footnotesize
\setlength{\tabcolsep}{5pt}
\begin{tabular}{@{}lrrr@{}}
\toprule
Calibration & Prefix & Acc.\ (\%) & Tokens \\
\midrule
\multicolumn{4}{@{}l}{\textit{MATH500}} \\
Block MSE & 8K & 77.2 & 8,862 \\
\method{} & 8K & 78.2 & 7,816 \\
\addlinespace[2pt]
Block MSE & 12K & 77.8 & 7,120 \\
\method{} & 12K & 79.4 & 6,697 \\
\midrule
\multicolumn{4}{@{}l}{\textit{AIME-120}} \\
Block MSE & 8K & 14.4 & 24,234 \\
\method{} & 8K & 16.7 & 22,279 \\
\addlinespace[2pt]
Block MSE & 12K & 15.0 & 20,866 \\
\method{} & 12K & 18.3 & 19,865 \\
\bottomrule
\end{tabular}\par\nointerlineskip
\end{minipage}\hfill
\begin{minipage}[t]{0.51\linewidth}
\vspace{0pt}
\centering
\includegraphics[width=\linewidth]{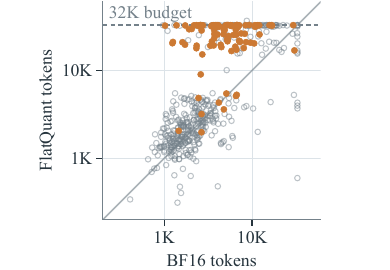}\par\nointerlineskip
\end{minipage}\par\nointerlineskip
\begin{minipage}[t]{0.45\linewidth}
\vspace{0pt}
\caption{Prefix and objective ablations on MATH500 and AIME-120 (R1-1.5B, W4A4KV4). Prefix: maximum history before the 4K block; Tokens: mean generation length.}
\label{tab:ablation-summary}
\end{minipage}\hfill
\begin{minipage}[t]{0.51\linewidth}
\vspace{0pt}
\captionof{figure}[Generation length and correctness under quantization.]{Paired MATH500 generation lengths for DeepSeek-R1-Distill-Qwen-1.5B: BF16 vs.\ FlatQuant W4A4KV4. Orange marks correct-to-incorrect cases; open gray points mark all others.}
\label{fig:generation-behavior}
\end{minipage}
\end{table}

\FloatBarrier
\section{Related Work}
\label{sec:related-work}

\paragraph{Post-training quantization of LLMs.}
LLM PTQ combines sensitivity-aware quantization, outlier suppression, and reconstruction-based calibration. GPTQ~\citep{frantar2023gptq} uses approximate second-order information for weight quantization, while AWQ~\citep{lin2024awq} protects salient channels through activation-aware scaling. SmoothQuant~\citep{xiao2023smoothquant} redistributes quantization difficulty from activations to weights, while QuaRot~\citep{ashkboos2024quarot} and SpinQuant~\citep{liu2025spinquant} suppress outliers through fixed and learned rotations, respectively. For reconstruction-based calibration, OmniQuant~\citep{shao2024omniquant} learns equivalent transformations and weight clipping, and FlatQuant~\citep{sun2025flatquant} learns affine transformations through layer-output reconstruction. Building on FlatQuant, \method{} extends calibration along the sequence dimension, preserving long historical context while distributing reconstruction across short blocks.

\paragraph{KV cache quantization.}
KV cache quantization balances memory savings, attention quality, and decoding overhead~\citep{lin2025qserve}. KIVI~\citep{liu2024kivi} exploits key/value asymmetry with per-channel key and per-token value quantization, while KVQuant~\citep{hooper2024kvquant} combines sensitivity-weighted quantization with outlier isolation. Outlier-aware rotations further improve low-bit cache representations~\citep{su2025rotatekv}, and vector codebooks provide an alternative to scalar quantization~\citep{li2025commvq}. Beyond cache distributions, OSCAR~\citep{zhou2026oscar} uses attention-aware covariance statistics to construct rotations and clipping thresholds.

\paragraph{Quantization and calibration for reasoning models.}
For reasoning models, quantization sensitivity varies across model families and tasks~\citep{liu2025quantization}, motivating reasoning-aware training~\citep{okoshi2026towards,lee2026reqat}. In post-training compression, calibration quality depends on both data selection~\citep{williams2024calibration} and sequence length~\citep{oh2025beyond}. Self-calibration generates calibration text with the model itself, avoiding dependence on external corpora~\citep{williams2025selfcalibration}. AYOT~\citep{wang2026ayot} uses reasoning traces and final answers from the high-precision target model alongside calibration questions.

\section{Conclusion}
We presented \method{}, a post-training quantization method that calibrates low-bit reasoning models on the history they read. Dynamic-prefix calibration brings the long quantized history into short-block reconstruction, and prefix read geometry supervises historical cache errors by their effect on attention, both only during calibration. Across multiple reasoning models, benchmarks, and quantization settings, \method{} consistently improves on the baseline, with the largest gains on long-horizon reasoning, although the remaining gap to full precision varies across models.

\subsection*{Ethics statement}
This study does not involve human participants. Potential biases, inaccurate outputs, and misuse should be considered when applying language models. Performance on standard benchmarks does not guarantee reliability across all use cases. Deployment in sensitive settings requires additional assessment and appropriate human oversight.

\subsection*{Reproducibility statement}
We retain experiment configurations, evaluation outputs, and analysis scripts to support verification of the reported results. These records document the settings and procedures used for the experiments and analyses. Reproduction should account for differences in software environments and the variability introduced by stochastic evaluation.

\subsection*{AI use statement}
AI tools were used to assist with manuscript revision and language refinement, including improvements to wording, clarity, and organization. We also used the full-precision target models to generate reasoning trajectories for self-calibration. The authors are responsible for the scientific content, the interpretation of the results, and the accuracy and integrity of the final manuscript.

\clearpage
\appendix
\FloatBarrier
\section{Implementation and Calibration Details}
\label{app:config}
\paragraph{Calibration data and budgets.}
The R1-1.5B pool contains 1,000 FP trajectories from NuminaMath-1.5, split into 900 source-train and 100 source-validation records. Filtering removes malformed responses and normalized exact-match overlaps with MATH500 and GSM8K; trajectories are used regardless of answer correctness. Fit cohorts retain leading segments of length-eligible trajectories. Table~\ref{tab:budget} gives the two relevant calibration budgets. Their cohorts differ, so the prefix comparison also changes data exposure.

\begin{table}[htbp]
\caption{Calibration budgets. Both configurations use a 4K active block, optimizer batch 4, and microbatch 1. Each position is reconstructed five times at 8K prefix and eight times at 12K.}
\label{tab:budget}
\centering\small
\begin{tabular}{lrrrrr}
\toprule
Configuration & Max prefix & Total context & Fit samples & Epochs & Updates/layer \\
\midrule
Historical ablation & 8K & 12K & 128 & 15 & 480 \\
Default & 12K & 16K & 64 & 32 & 512 \\
\bottomrule
\end{tabular}
\end{table}

Transformation and diagonal parameters use the base learning rate in Section~\ref{sec:setup}; learned clipping uses a tenfold multiplier. Gradients are accumulated with detached loss normalization. Both gradient norms used to set $\alpha_\ell$ cover the same trainable parameters, with unused gradients treated as zero, and are measured without updating the initialization. The default recipe collects geometry from the original FP model. Calibration preserves causal order and original position IDs. At every microstep, prefix K/V are regenerated from fixed FP layer inputs using the current quantization parameters, with gradients retained through prefix construction.

\paragraph{Query sampling.}
\label{app:curvature}
For each sample and nonempty prefix, divide eligible assistant queries in the active block into four position strata and sample up to 16 queries without replacement per stratum. A sampled query from stratum $a$ receives weight $N_a/(n_a|\mathcal Q_P|)$, where $N_a$ and $n_a$ are its eligible and sampled counts. These inverse-inclusion weights estimate the mean in Equation~\ref{eq:read-blocks}. Query selections remain fixed across layers and GQA heads.

\paragraph{Shared key directions.}
Let $\widehat G^K_{i,t,P,h}$ be a sampled key block and $s^K_{i,t,P,h}$ its trace. For each 4K position bucket $\beta$ and KV head $h$, pool positive-trace blocks across fit samples and prefix conditions:
\begin{equation}
\overline G^K_{\beta,h}
=\frac{1}{|\mathcal I_{\beta,h}|}
\sum_{(i,t,P)\in\mathcal I_{\beta,h}}
\frac{\widehat G^K_{i,t,P,h}}{s^K_{i,t,P,h}}.
\end{equation}
Entry strengths remain sample- and prefix-specific. If $\lambda_1\geq\cdots\geq\lambda_{d_h}$ are the pooled eigenvalues and $U_r$ contains the leading eigenvectors, use $r=8$ and
\begin{equation}
\rho=\frac{1}{d_h-r}\sum_{j=r+1}^{d_h}\lambda_j,\qquad
\widetilde G^K=\rho I+
U_r\operatorname{diag}(\lambda_1-\rho,\ldots,\lambda_r-\rho)U_r^\top.
\label{eq:lowrank-read}
\end{equation}
This PSD approximation preserves the pooled trace and replaces discarded directions with an isotropic residual. Values retain scalar read strengths. Cache errors use the current true right inverses with gradients, as in Section~\ref{sec:curvature}.

\paragraph{Prefix computation and cost.}
Active-block outputs match the corresponding full-forward outputs when the layer inputs, cache transformations, original positions, tokenwise quantization statistics, and causal support are identical and prefix gradients are retained. Each update computes $BP+B(B+1)/2$ query--key pairs, compared with $T(T+1)/2$ for full-sequence reconstruction; with $T=4B$, this is about a quarter of the full-sequence pairs on average over the four blocks, and feed-forward outputs are computed only for the $B$ active positions. Prefix K/V are still projected at every update, the rank-$r$ penalty adds $O(PH_{\mathrm{KV}}d_hr)$ work, and read-geometry statistics are collected once per model before calibration.

\clearpage
\section{Evidence for the Motivation}
\label{app:diagnostics}
\paragraph{Transfer of importance scores.}
\label{app:transfer}
We label the top 10\% of cache repairs by their reduction in teacher-forced answer loss. Each of four folds fits scores on three trajectories (1,536 candidates) and evaluates on the held-out trajectory (512 candidates); clusters are fitted on training trajectories only. Table~\ref{tab:transfer} reports all five tested scores. None exceeds an AUC of 0.51, including the future-read count, which uses information unavailable during calibration. These are four discovery trajectories from an earlier V-codeword configuration, so the results support the limited-transfer observation for these scores, not a general impossibility claim or a matched W4A4KV4 comparison.

\begin{table}[htbp]
\caption{Importance-score transfer, averaged over four held-out trajectories. AP is average precision; P@10\% is precision among the selected 10\%. Future-read count uses information unavailable during calibration and is computed from a separate trace; it serves as a reference.}
\label{tab:transfer}
\centering\small
\begin{tabular}{lrrr}
\toprule
Score & AUC & AP & P@10\% \\
\midrule
Codeword shrinkage & 0.495 & 0.109 & 0.096 \\
Residual-direction cluster & 0.504 & 0.108 & 0.111 \\
Position-bin shrinkage & 0.429 & 0.091 & 0.053 \\
Future-read count & 0.495 & 0.107 & 0.091 \\
Local V error & 0.504 & 0.103 & 0.067 \\
\bottomrule
\end{tabular}
\end{table}

\paragraph{Position coverage.}
\label{app:position}
The 8K diagnostic evaluates eight development trajectories in their original causal context, averaging FP-to-quantized KL within 1K position buckets. Short-window and fixed-tail calibration fail in opposite regions, whereas full 8K calibration stays between 0.140 and 0.190 nats/token in every bucket. Rotating-prefix calibration over a 12K context, evaluated on four development trajectories, likewise stays between 0.101 and 0.191 nats/token, so no region of the sequence is left unsupervised. Different cohorts prevent direct magnitude comparisons between these diagnostics; they do not measure task accuracy or isolate rotation from all other calibration changes.

\end{document}